\documentclass[11pt,a4paper,draft]{article}
\usepackage[hyperref]{emnlp2018}
\usepackage{times}
\usepackage{latexsym}

\usepackage{url}

\usepackage{amsmath}
\usepackage{booktabs}
\usepackage{comment}
\usepackage{hhline}
\usepackage[textsize=small]{todonotes}
\usepackage[toc,page]{appendix}

\aclfinalcopy 

\title{Augmenting Statistical Machine Translation with \\ Subword Translation of Out-of-Vocabulary Words}

\author{ 
    Nelson F. Liu$^{\spadesuit\diamondsuit}$ \quad
	{\bf Jonathan May}$^\clubsuit$ \quad
	{\bf Michael Pust}$^\clubsuit$ \quad 
	{\bf Kevin Knight}$^\clubsuit$ \\
    $^\spadesuit$Paul G. Allen School of Computer Science \& Engineering, University of Washington\\
	$^\diamondsuit$Department of Linguistics, University of Washington \\
    	{\tt nfliu@cs.washington.edu} \\
	$^\clubsuit$Information Sciences Institute, University of Southern California \\
	{\tt \{jonmay,pust,knight\}@isi.edu}
}
\date{}

\begin{document}
\maketitle
\begin{abstract}
Most statistical machine translation systems cannot translate words that are unseen in the training data. 
However, humans can translate many classes of out-of-vocabulary (OOV) words (e.g., novel morphological variants, misspellings, and compounds) without context by using orthographic clues. 
Following this observation, we describe and evaluate several general methods for OOV translation that use only subword information.
We pose the OOV translation problem as a standalone task and intrinsically evaluate our approaches on fourteen typologically diverse languages across varying resource levels. 
Adding OOV translators to a statistical machine translation system yields consistent BLEU gains (0.5 points on average, and up to 2.0) for all fourteen languages, especially in low-resource scenarios.
\end{abstract}

\section{Introduction}
Machine translation systems frequently must translate tokens unseen during training (known as out-of-vocabulary or OOV tokens). 
Neural machine translation (NMT) can mitigate this OOV problem by producing word representations on the fly from subword units \cite{Sennrich2016NeuralMT,Luong2016AchievingOV}.
Despite this advantage, NMT performs poorly when there is little training data; \citet{Koehn2017SixCN} show that statistical machine translation (SMT) yields better translations in this low resource setting. 
However, SMT systems struggle to handle OOV tokens.

Human translators are adept at translating OOV words, in part because they exploit subword orthographic clues. 
For example, they can translate novel morphological variants and compounds of known words by reasoning over constituent subword units. We use these same subword orthographic units to build broadly-applicable OOV translation systems while making only the loose typological assumption that the orthographic representation contains informative subword units.

Most prior work has focused on translating specific OOV classes; we pursue holistic solutions. 
Our work is similar in spirit to prior language-specific and general methods for handling multiple classes of OOVs in SMT \citep{habash2009remoov, Gujral2016TranslatingUW}.
We compare three approaches to building a subword OOV translation module: an \textbf{edit distance} approach, which matches OOV words to orthographically similar known translation pairs; a \textbf{vector distance} approach, which seeks a semantic match instead of a orthographic match; and a \textbf{sequence-to-sequence} (seq2seq) approach, which generates translations one character at a time. 
We do not use language-specific heuristics, enabling automatic construction of OOV modules for a wide variety of languages.

We evaluate our approaches on an intrinsic OOV translation task on a typologically diverse set of fourteen languages.
We also embed OOV translators into a syntax-based machine translation (SBMT) system and assess its effects on overall system BLEU for the same fourteen languages.
Our results show that using an OOV translator with the SBMT system consistently improves translation quality across all languages, especially in low resource scenarios; we see gains of 0.5 BLEU on average and up to 2.0 BLEU. 
We release code to train our OOV translators at \url{http://nelsonliu.me/papers/oov}.

\section{Dataset and Intrinsic Evaluation}

To intrinsically evaluate OOV modules, we assess their ability to translate previously unseen foreign tokens into English. 
For fourteen language pairs, we obtain monolingual data, sentence-aligned parallel data, and bilingual lexicons. 
We word-align \cite{Och2003ASC, Liang-Taskar-Klein:2006:Alignment} the parallel data and randomly remove 1000 one-count \texttt{<foreign, English>} word pairs that also do not exist in the lexicons. 
These word translation pairs are out-of-vocabulary {\em with respect to the other resources}, making them suitable for intrinsic evaluation.
We divide these pairs into validation and test splits of 500 word pairs each, and build our OOV translators with the monolingual data, the lexicons, and translation tables extracted from the parallel text.

To summarize, our dataset contains: 
(1) {\bf Validation and test sets}: Foreign OOV tokens and an English translation\footnote{While there are often multiple acceptable English translations for a foreign OOV, our dataset provides one.} (all extracted from parallel text). 
Our objective is to predict the English translation, given the foreign OOV. 
(2) {\bf Lexicon} (bilingual dictionary): Foreign tokens, their part of speech, and an English translation. 
A foreign token may have multiple entries. 
(3) {\bf Monolingual data}: A modest amount of running text in the foreign language (from, e.g., Wikipedia).
(4) {\bf Translation tables}: Pairs of aligned foreign words and their English translations (token to token mapping) with alignment probabilities and absolute alignment counts, derived from the parallel text. 
For dataset quality validation details, see Appendix~\ref{appendix:dataset_validation}.

\section{OOV Translation Methods}

\paragraph{Edit Distance}
To translate OOVs with edit distance \cite{Levenshtein1966}, we adapt the method of \citet{Gujral2016TranslatingUW}. 
We begin by retrieving the foreign word(s) in the bilingual lexicon or translation table with the lowest edit distance from the given OOV token. 
Our predicted translation is the English word that most frequently\footnote{Ties are broken with the words' frequency in the Gigaword corpus.} aligns to any of the selected in-vocabulary foreign words. 

\paragraph{Vector Distance} To use vector distance for OOV translation, we calculate the cosine similarity between word vectors to select the in-vocabulary word with the closest word vector to the OOV. 

Our predicted translation is again the English word that most frequently aligns to the selected in-vocabulary source token.

To obtain vectors for OOVs, we use FastText models \cite{Bojanowski2017EnrichingWV} trained on source language monolingual data. 
Since FastText computes vectors from subword units, it can produce representations for arbitrary strings; we thus use FastText vectors for both the input OOV tokens and the in-vocabulary words.

Prior work has used word vectors for handling OOVs \citep[and more]{zou2013bilingual, zhang2014bilingually,Madhyastha2017LearningBP}, but the majority learn a bilingual mapping between the source and target languages. 
Our method does not learn such a mapping, reducing our reliance on parallel data. 
Using subword vectors enables translation of OOVs unseen in the monolingual data.

\paragraph{Sequence-to-Sequence} The edit and vector distance methods are constrained to only output translations that occur in the bilingual dictionary or the translation table. 
Towards open-vocabulary OOV translation, we use character level sequence-to-sequence (seq2seq) \citep{SutskeverSeq2Seq} models to generate English translations from source strings. 
This approach is similar in spirit to prior work on word-level NMT models that back off to character-level information for OOV tokens \citep{Luong2016AchievingOV}. 
To the best of our knowledge, this is the first use of seq2seq for translating OOVs in SMT.

We use an LSTM-based seq2seq model with attention, which is trained on source-target pairs extracted from the translation table and bilingual dictionary. 
We weight the examples in our training data, since certain translations are more common than others. 
Pairs extracted from the translation table are weighted by their absolute alignment frequency, and pairs from the bilingual dictionary are given a constant weight of 100. 
See Appendices~\ref{appendix:seq2seq_oov_data_size}~and~\ref{appendix:seq2seq_oov_details} for training dataset sizes and implementation details.

\section{Experiments and Results}

To intrinsically evaluate OOV module performance, we measure the proportion of predicted translations that exactly match target translations.

In addition, we measure the effect of integrating OOV translation systems into an SBMT system. 
We incorporate our OOV translation systems into an end-to-end machine translation system by adding OOVs and their predicted translations as translation pairs (for syntax-based MT, part-of-speech-tagged translation pairs) with an indicator feature that is tuned with other standard features. 
These pairs compete with do-not-translate pairs (i.e. where the source and target are identical); feature weights and language model scores determine whether the system uses a translated OOV or chooses to not translate it. 

\subsection{Intrinsic OOV Translation Results}

The accuracy of each OOV translation method on each of the fourteen languages is presented in Table~\ref{tab:intrinsic_test}. 
On average, the seq2seq models outperform the edit distance systems, followed by the vector distance OOV translation systems.

\begin{table}[!htp]
\footnotesize
\centering
\begin{tabular}{lrrr}
\toprule
\begin{tabular}[c]{@{}c@{}}Source\\ Language\end{tabular} & \multicolumn{1}{c}{\begin{tabular}[c]{@{}c@{}}Edit \\ Distance\end{tabular}} & \multicolumn{1}{c}{\begin{tabular}[c]{@{}c@{}}Vector \\ Distance\end{tabular}} & \multicolumn{1}{c}{Seq2Seq} \\
\midrule
Amharic & 22.8\% & 14.2\% & \textbf{27.0\%} \tabularnewline
Arabic & 20.0\% & 15.8\% & \textbf{29.4\%} \tabularnewline
Bengali & \textbf{23.5\%} & 23.2\% & 20.5\% \tabularnewline
Farsi & 27.2\% & 25.2\% & \textbf{35.6\%} \tabularnewline
Hausa & 23.0\% & 7.2\% & \textbf{24.8\%} \tabularnewline
Hungarian & 23.4\% & 19.6\% & \textbf{32.4\%} \tabularnewline
Russian & 20.0\% & 20.2\% & \textbf{33.4\%} \tabularnewline
Somali & 30.4\% & 18.4\% & \textbf{37.2\%} \tabularnewline
Spanish & 20.8\% & 16.8\% & \textbf{28.6\%} \tabularnewline
Tamil & 21.9\% & 21.4\% & \textbf{28.7\%} \tabularnewline
Turkish & 29.2\% & 28.8\% & \textbf{38.6\%} \tabularnewline
Urdu & 13.3\% & \textbf{17.4\%} & 10.2\% \tabularnewline
Uzbek & 22.6\% & 21.2\% & \textbf{36.4\%} \tabularnewline
Yoruba & 14.6\% & 11.0\% & \textbf{19.8\%} \tabularnewline
\midrule
Average & 22.34\% & 19.31\% & \textbf{28.76}\% \tabularnewline
\bottomrule
\end{tabular}
\caption{\label{tab:intrinsic_test} Intrinsic test set exact match accuracy for each of the translation methods. For all source languages, the target is English. Bold marks the best performing method for each pair.}
\end{table} 

\subsection{Extrinsic SBMT Results}

\begin{table*}[!htp]
\footnotesize
\centering
\setlength\tabcolsep{2.5pt} 
\begin{tabular}{rrrrrrrrrrrrrrrrr}
\toprule
\multicolumn{16}{c}{Source Language Code} \tabularnewline
& \multicolumn{1}{r}{avg} & \multicolumn{1}{r}{avg $\Delta$} & \multicolumn{1}{r}{amh} & \multicolumn{1}{r}{ara} & \multicolumn{1}{r}{ben} & \multicolumn{1}{r}{fas} & \multicolumn{1}{r}{hau} & \multicolumn{1}{r}{hun} & \multicolumn{1}{r}{rus} & \multicolumn{1}{r}{som} & \multicolumn{1}{r}{spa} & \multicolumn{1}{r}{tam} & \multicolumn{1}{r}{tur} & \multicolumn{1}{r}{urd} & \multicolumn{1}{r}{uzb} & \multicolumn{1}{r}{yor} \tabularnewline
 \midrule
SBMT & 21.36 & - & 15.75 & 21.13 & 10.92 & 24.12 & 21.81 & 17.56 & 31.36 & 21.96 & 40.36 & 20.77 & 20.12 & 18.22 & 16.86 & 18.09 \tabularnewline
edit dist. & 21.72 & +0.36 & 15.76 & 22.81 & \textbf{11.31} &  \textbf{25.22} & \textbf{22.20} & \textbf{18.39} & 31.45 & \textbf{22.63} & 40.94 & 21.98 & 17.29 & \textbf{18.86} & 17.13 & 18.06 \tabularnewline
vector dist. & 21.61 & +0.25 & \textbf{16.07} & 23.02 & 10.16 &  23.87 & 21.55 & 17.78 & 31.52 & 22.19 & \textbf{41.00} & \textbf{22.58} & 20.30 & 18.08 & \textbf{17.39} & 17.06 \tabularnewline
seq2seq & \textbf{21.86} & \textbf{+0.50} & 15.84 & \textbf{23.17} & 10.92 & 24.44 & 21.85 & 17.74 & \textbf{32.04} & \textbf{22.63} & 40.73 & 22.35 & \textbf{20.49} & 18.49 & 17.07 & \textbf{18.30} \tabularnewline
\midrule
BPE NMT & 10.72 & -10.71 & 6.85 & 8.92 & 3.74 & 15.08 & 15.63 & 6.51 & 8.61 & 13.02 & 20.31 & 10.67 & 6.93 & 12.8 & 8.91 & 12.12 \tabularnewline
\bottomrule
\end{tabular}
\caption{Test SBMT BLEU scores for each language pair and OOV translation method. The best OOV translation method for each pair is bolded. BLEU scores of BPE NMT trained on same data are also provided for comparison.}
\label{tab:extrinsic_test}
\end{table*}

Table~\ref{tab:extrinsic_test} illustrates the effects of our OOV module on SBMT BLEU across the fourteen languages. 
We compare against a baseline subword NMT system trained on the same data with source and target-side byte pair encoding (BPE; \citealp{Sennrich2016NeuralMT}).\footnote{For NMT baseline implementation details, see Appendix~\ref{appendix:nmt_baseline_details}} 
All MT systems are trained on between 262K to 11.9M words; see Appendix~\ref{appendix:sbmt_nmt_baselines_data_size} for the amount of training data per language.

On average, adding the seq2seq OOV translator to SBMT produced the highest BLEU scores, followed by the edit and vector distance methods.
Notably, the seq2seq OOV translator improves SBMT BLEU for all languages except Bengali, where it ties with the vanilla SBMT baseline.
For each language, at least one of the OOV-augmented systems improves upon the SBMT baseline. 
This confirms that OOV translation from subword information has broad utility; adding an OOV translator to SBMT is an easy and consistent way to improve performance.
We see average gains of 0.5 BLEU points, with a 2.0 BLEU improvement for Arabic.
SBMT with or without OOV translation outperforms the BPE NMT models, supporting previous observations that SMT is superior in low resource scenarios.\footnote{For reference, \citet{Koehn2017SixCN} report that NMT begins to outperform SMT for English-Spanish when trained on more than approximately 15 million words.} 
This also further motivates OOV translation in SMT, since directly applying subword NMT is clearly impractical here.

\section{Discussion}
\paragraph{Method Performance by OOV Type} To further study the ability of our OOV translation methods to handle various types of OOV tokens, we randomly sampled 100 examples from the development set used in the Spanish-English intrinsic OOV translation task and broadly categorized OOVs by whether they are morphological variations of an in-vocabulary word, misspellings, a transliteration, a compound word, or whether the OOV is a proper noun that should be copied to the translation.
The performance of each method for each category is presented in Table~\ref{tab:stratified_accuracy}. 

The seq2seq methods show the best performance on examples involving morphological variation, since they reason over subword units. This also explains the large BLEU gains when adding seq2seq OOV translation to Arabic SBMT, as the language is morphologically complex and many OOVs are morphological variants of known words.
Reasoning over subword units also enables seq2seq translators to handle OOV tokens created from compounding, where the edit and vector distance methods struggle. 
The edit distance-based method intuitively outpaces the others on OOV words generated by misspellings, and draws even with the seq2seq methods on transliteration cases.

Many of the proper nouns are rare words, which the edit and vector distance methods cannot handle.
The seq2seq model performs slightly better.

\begin{table*}[htpb]
\footnotesize
\centering
\begin{tabular}{lrrrrr}
\toprule
OOV Category & Occurrences in Sample & Edit Distance & Vector Distance & Seq2Seq \\ 
\midrule
Morphological Variation & 57 & 14.0\% & 10.5\% & \textbf{24.6\%} \tabularnewline
Misspelling & 19 & \textbf{31.6\%} & 26.3\% & 21.1\% \tabularnewline
Transliteration & 14 & \textbf{21.4\%} & 14.3\% & \textbf{21.4\%} \tabularnewline
Compounding & 5 & 0.0\% & 0.0\% & \textbf{40.0\%} \tabularnewline
Proper Noun & 5 & 0.0\% & 20.0\% & \textbf{60.0\%} \tabularnewline
\midrule
All & 100 & 17.0\% & 14.0\% & \textbf{26.0\%} \tabularnewline
\bottomrule
\end{tabular}
\caption{\label{tab:stratified_accuracy} Exact-match accuracy of each OOV translation method, stratified by OOV type. The best OOV translation method for each category is bolded.}
\end{table*}

\paragraph{Seq2Seq Produces English Words Unseen During Training} 

Seq2seq models are able to compose units of meaning to produce novel target-side tokens unseen during training; we see this theoretical advantage in practice.
In the examples in Table~\ref{tab:examples}, the seq2seq-predicted translations did not occur in the target side of the training data, so the model must have combined subword units to produce them. The model learns to (a) combine previously seen subword units into novel English compounds, (b) transliterate sequences, and (c) inflect verbs for which it has seen a root form.

\begin{table}[!hptb]
\footnotesize
\begin{tabular}{ll}
\multicolumn{2}{l}{\textbf{(a) Compounding}} \\ \hhline{=|=}
\multicolumn{1}{l|}{SPA source OOV} & ciberviolencia \\ \hline
\multicolumn{1}{l|}{ENG gold translation} & cyberviolence \\ \hline
\multicolumn{1}{l|}{edit distance prediction} & Roscomnadzor \\ \hline
\multicolumn{1}{l|}{seq2seq prediction} & cyberviolence \\ \hline
\end{tabular}

\bigskip

\begin{tabular}{ll}
\multicolumn{2}{l}{\textbf{(b) Transliteration / Copying}} \\ \hhline{=|=}
\multicolumn{1}{l|}{SPA source OOV} & Kafkast\'{a}n \\ \hline
\multicolumn{1}{l|}{ENG gold translation} & Kafkastan \\ \hline
\multicolumn{1}{l|}{edit distance prediction} & alternative form of kazajist\'{a}n \\ \hline
\multicolumn{1}{l|}{seq2seq prediction} & Kafkastan \\ \hline
\end{tabular}

\bigskip

\begin{tabular}{ll}
\multicolumn{2}{l}{\textbf{(c) Morphology}} \\ \hhline{=|=}
\multicolumn{1}{l|}{SPA source OOV} & balcanizada \\ \hline
\multicolumn{1}{l|}{ENG gold translation} & balkanised \\ \hline
\multicolumn{1}{l|}{edit distance prediction} & unbanked \\ \hline
\multicolumn{1}{l|}{seq2seq prediction} & balkanized \\ \hline
\end{tabular}

\caption{\label{tab:examples} seq2seq models recombine in-vocabulary tokens to output novel words unseen during training.}
\end{table}

\section{Related Work}
Many strategies have been developed for OOV translation, especially in SMT. For example, \citet{Nieen2000ImprovingSQ, Koehn2003EmpiricalMF, Virpioja2007MorphologyAwareSM} translate OOV compounds and other morphologically complex words by splitting and translating the resultant segments.
\citet{AlOnaizan2002MachineTO, Habash2008FourTF, Hermjakob2008NameTI, Durrani2014IntegratingAU} explore transliteration for OOV named entities.

Many approaches also translate OOV tokens by expanding the translation lexicon with additional bilingual or monolingual resources \citep[among others]{rapp1995identifying, callison2006improved, haghighi2008learning, marton2009improved, daume11lexicaladapt, razmara2013graph, Irvine2013CombiningBA, Mikolov2013ExploitingSA, saluja2014graph, zhao2015learning}. 

OOV translation has also been cast as a problem of decipherment \citep{ravi2011deciphering,dou2012large}, and other approaches use information from cognates or related languages  \citep[among others]{hajivc2000machine,kondrak2003cognates,de2006catalan,durrani2010hindi,wang2012source,nakov2012improving,dholakia2014pivot,tsvetkov2015lexicon}.

\section{Conclusion}

We compare three generally-applicable strategies for translating out-of-vocabulary words, none of which rely on any language-specific resources or typological assumptions beyond the presence of subword units.
Integrating these OOV translators into a SMT system consistently improves translation quality over a typologically diverse set of of fourteen languages. We analyze method performance over a range of OOV types and also demonstrate that seq2seq OOV translators compose characters to generate novel target-side translations.
\bibliography{oov_translation}

\begin{thebibliography}{43}
\expandafter\ifx\csname natexlab\endcsname\relax\def\natexlab#1{#1}\fi

\bibitem[{Al-Onaizan and Knight(2002)}]{AlOnaizan2002MachineTO}
Yaser Al-Onaizan and Kevin Knight. 2002.
\newblock Machine {T}ransliteration of {N}ames in {A}rabic {T}ext.
\newblock In \emph{Proc. of the ACL Workshop on Computational Approaches to
  Semitic Languages}, pages 1--13.

\bibitem[{Bojanowski et~al.(2017)Bojanowski, Grave, Joulin, and
  Mikolov}]{Bojanowski2017EnrichingWV}
Piotr Bojanowski, Edouard Grave, Armand Joulin, and Tomas Mikolov. 2017.
\newblock {E}nriching {W}ord {V}ectors with {S}ubword {I}nformation.
\newblock \emph{Transactions of the ACL}, 5:135--146.

\bibitem[{Callison-Burch et~al.(2006)Callison-Burch, Koehn, and
  Osborne}]{callison2006improved}
Chris Callison-Burch, Philipp Koehn, and Miles Osborne. 2006.
\newblock Improved {S}tatistical {M}achine {T}ranslation {U}sing {P}araphrases.
\newblock In \emph{Proc. of NAACL}, pages 17--24.

\bibitem[{{Daum\'e III} and Jagarlamudi(2011)}]{daume11lexicaladapt}
Hal {Daum\'e III} and Jagadeesh Jagarlamudi. 2011.
\newblock {D}omain {A}daptation for {M}achine {T}ranslation by {M}ining
  {U}nseen {W}ords.
\newblock In \emph{Proc. of NAACL}, pages 407--412.

\bibitem[{De~Gispert and Marino(2006)}]{de2006catalan}
Adri{\`a} De~Gispert and Jose~B Marino. 2006.
\newblock {C}atalan-{E}nglish {S}tatistical {M}achine {T}ranslation without
  {P}arallel {C}orpus: {B}ridging through {S}panish.
\newblock In \emph{Proc. of LREC}, pages 65--68.

\bibitem[{Dholakia and Sarkar(2014)}]{dholakia2014pivot}
Rohit Dholakia and Anoop Sarkar. 2014.
\newblock {P}ivot-based {T}riangulation for {L}ow-{R}esource {L}anguages.
\newblock In \emph{Proc. of AMTA}, pages 315--328.

\bibitem[{Dou and Knight(2012)}]{dou2012large}
Qing Dou and Kevin Knight. 2012.
\newblock {L}arge {S}cale {D}ecipherment for {O}ut-of-{D}omain {M}achine
  {T}ranslation.
\newblock In \emph{Proc. of EMNLP}, pages 266--275.

\bibitem[{Durrani et~al.(2010)Durrani, Sajjad, Fraser, and
  Schmid}]{durrani2010hindi}
Nadir Durrani, Hassan Sajjad, Alexander Fraser, and Helmut Schmid. 2010.
\newblock Hindi-to-{U}rdu {M}achine {T}ranslation through {T}ransliteration.
\newblock In \emph{Proc. of ACL}, pages 465--474.

\bibitem[{Durrani et~al.(2014)Durrani, Sajjad, Hoang, and
  Koehn}]{Durrani2014IntegratingAU}
Nadir Durrani, Hassan Sajjad, Hieu Hoang, and Philipp Koehn. 2014.
\newblock {I}ntegrating an {U}nsupervised {T}ransliteration {M}odel into
  {S}tatistical {M}achine {T}ranslation.
\newblock In \emph{Proc. of ACL}, pages 148--153.

\bibitem[{Gujral et~al.(2016)Gujral, Khayrallah, and
  Koehn}]{Gujral2016TranslatingUW}
Biman Gujral, Huda Khayrallah, and Philipp Koehn. 2016.
\newblock {T}ranslation of {U}nknown {W}ords in {L}ow {R}esource {L}anguages.
\newblock In \emph{Proc. of AMTA}.

\bibitem[{Habash(2008)}]{Habash2008FourTF}
Nizar Habash. 2008.
\newblock {F}our {T}echniques for {O}nline {H}andling of {O}ut-of-{V}ocabulary
  {W}ords in {A}rabic-{E}nglish {S}tatistical {M}achine {T}ranslation.
\newblock In \emph{Proc. of ACL}, pages 57--60.

\bibitem[{Habash(2009)}]{habash2009remoov}
Nizar Habash. 2009.
\newblock {REMOOV}: {A} {T}ool for {O}nline {H}andling of {O}ut-of-{V}ocabulary
  {W}ords in {M}achine {T}ranslation.
\newblock In \emph{Proc. of MEDAR}.

\bibitem[{Haghighi et~al.(2008)Haghighi, Liang, Berg-Kirkpatrick, and
  Klein}]{haghighi2008learning}
Aria Haghighi, Percy Liang, Taylor Berg-Kirkpatrick, and Dan Klein. 2008.
\newblock Learning {B}ilingual {L}exicons from {M}onolingual {C}orpora.
\newblock \emph{Proc. of ACL}, pages 771--779.

\bibitem[{Haji{\v{c}} et~al.(2000)Haji{\v{c}}, Hric, and
  Kubo{\v{n}}}]{hajivc2000machine}
Jan Haji{\v{c}}, Jan Hric, and Vladislav Kubo{\v{n}}. 2000.
\newblock {M}achine {T}ranslation of {V}ery {C}lose {L}anguages.
\newblock In \emph{Proc. of ANLP}, pages 7--12.

\bibitem[{Hermjakob et~al.(2008)Hermjakob, Knight, and
  Daum{\'e}~III}]{Hermjakob2008NameTI}
Ulf Hermjakob, Kevin Knight, and Hal Daum{\'e}~III. 2008.
\newblock {N}ame {T}ranslation in {S}tatistical {M}achine {T}ranslation -
  {L}earning {W}hen to {T}ransliterate.
\newblock In \emph{Proc. of ACL}, pages 389--397.

\bibitem[{Irvine and Callison-Burch(2013)}]{Irvine2013CombiningBA}
Ann Irvine and Chris Callison-Burch. 2013.
\newblock {C}ombining {B}ilingual and {C}omparable {C}orpora for {L}ow
  {R}esource {M}achine {T}ranslation.
\newblock In \emph{Proc. of the Eighth Workshop on Statistical Machine
  Translation}, pages 262--270.

\bibitem[{Kingma and Ba(2014)}]{Kingma2014AdamAM}
Diederik~P. Kingma and Jimmy Ba. 2014.
\newblock {A}dam: {A} {M}ethod for {S}tochastic {O}ptimization.
\newblock ArXiv:1412.6980.

\bibitem[{Klein et~al.(2017)Klein, Kim, Deng, Senellart, and Rush}]{opennmt}
Guillaume Klein, Yoon Kim, Yuntian Deng, Jean Senellart, and Alexander~M. Rush.
  2017.
\newblock Open{NMT}: {O}pen-{S}ource {T}oolkit for {N}eural {M}achine
  {T}ranslation.
\newblock In \emph{Proc. of ACL}.

\bibitem[{Koehn and Knight(2003)}]{Koehn2003EmpiricalMF}
Philipp Koehn and Kevin Knight. 2003.
\newblock {E}mpirical {M}ethods for {C}ompound {S}plitting.
\newblock In \emph{Proc. of EACL}, pages 187--193.

\bibitem[{Koehn and Knowles(2017)}]{Koehn2017SixCN}
Philipp Koehn and Rebecca Knowles. 2017.
\newblock {S}ix {C}hallenges for {N}eural {M}achine {T}ranslation.
\newblock In \emph{Proc. of the First Workshop on Neural Machine Translation},
  pages 28--39.

\bibitem[{Kondrak et~al.(2003)Kondrak, Marcu, and Knight}]{kondrak2003cognates}
Grzegorz Kondrak, Daniel Marcu, and Kevin Knight. 2003.
\newblock {C}ognates {C}an {I}mprove {S}tatistical {T}ranslation {M}odels.
\newblock In \emph{Proc. of NAACL}, pages 46--48.

\bibitem[{{Levenshtein}(1966)}]{Levenshtein1966}
V.~I. {Levenshtein}. 1966.
\newblock {B}inary {C}odes {C}apable of {C}orrecting {D}eletions, {I}nsertions
  and {R}eversals.
\newblock \emph{Soviet Physics Doklady}, 10:707.

\bibitem[{Liang et~al.(2006)Liang, Taskar, and
  Klein}]{Liang-Taskar-Klein:2006:Alignment}
Percy Liang, Ben Taskar, and Dan Klein. 2006.
\newblock {A}lignment by {A}greement.
\newblock In \emph{Proc. of NAACL}, pages 104--111.

\bibitem[{Luong and Manning(2016)}]{Luong2016AchievingOV}
Minh-Thang Luong and Christopher~D. Manning. 2016.
\newblock {A}chieving {O}pen {V}ocabulary {N}eural {M}achine {T}ranslation with
  {H}ybrid {W}ord-{C}haracter {M}odels.
\newblock In \emph{Proc. of ACL}, pages 1054--1063.

\bibitem[{Luong et~al.(2015)Luong, Pham, and Manning}]{Luong2015EffectiveAT}
Thang Luong, Hieu Pham, and Christopher~D. Manning. 2015.
\newblock {E}ffective {A}pproaches to {A}ttention-based {N}eural {M}achine
  {T}ranslation.
\newblock In \emph{Proc. of EMNLP}, pages 1412--1421.

\bibitem[{Madhyastha and Espa\~{n}a Bonet(2017)}]{Madhyastha2017LearningBP}
Pranava~Swaroop Madhyastha and Cristina Espa\~{n}a Bonet. 2017.
\newblock {L}earning {B}ilingual {P}rojections of {E}mbeddings for {V}ocabulary
  {E}xpansion in {M}achine {T}ranslation.
\newblock In \emph{Proc. of the 2nd Workshop on Representation Learning for
  NLP}, pages 139--145.

\bibitem[{Marton et~al.(2009)Marton, Callison-Burch, and
  Resnik}]{marton2009improved}
Yuval Marton, Chris Callison-Burch, and Philip Resnik. 2009.
\newblock {I}mproved {S}tatistical {M}achine {T}ranslation {U}sing
  {M}onolingually-{D}erived {P}araphrases.
\newblock In \emph{Proc. of EMNLP}, pages 381--390.

\bibitem[{Mikolov et~al.(2013)Mikolov, Le, and
  Sutskever}]{Mikolov2013ExploitingSA}
Tomas Mikolov, Quoc~V. Le, and Ilya Sutskever. 2013.
\newblock {E}xploiting {S}imilarities among {L}anguages for {M}achine
  {T}ranslation.
\newblock ArXiv:1309.4168.

\bibitem[{Nakov and Ng(2012)}]{nakov2012improving}
Preslav Nakov and Hwee~Tou Ng. 2012.
\newblock {I}mproving {S}tatistical {M}achine {T}ranslation for a
  {R}esource-poor {L}anguage {U}sing {R}elated {R}esource-{R}ich {L}anguages.
\newblock \emph{Journal of Artificial Intelligence Research}, 44:179--222.

\bibitem[{Nie{\ss}en and Ney(2000)}]{Nieen2000ImprovingSQ}
Sonja Nie{\ss}en and Hermann Ney. 2000.
\newblock {I}mproving {SMT} {Q}uality with {M}orpho-syntactic {A}nalysis.
\newblock In \emph{Proc. of COLING}, pages 1081--1085.

\bibitem[{Och and Ney(2003)}]{Och2003ASC}
Franz~Josef Och and Hermann Ney. 2003.
\newblock A {S}ystematic {C}omparison of {V}arious {S}tatistical {A}lignment
  {M}odels.
\newblock \emph{Computational Linguistics}, 29:19--51.

\bibitem[{Rapp(1995)}]{rapp1995identifying}
Reinhard Rapp. 1995.
\newblock {I}dentifying {W}ord {T}ranslations in {N}on-{P}arallel {T}exts.
\newblock In \emph{Proc. of ACL}, pages 320--322.

\bibitem[{Ravi and Knight(2011)}]{ravi2011deciphering}
Sujith Ravi and Kevin Knight. 2011.
\newblock Deciphering {F}oreign {L}anguage.
\newblock In \emph{Proc. of ACL}, pages 12--21.

\bibitem[{Razmara et~al.(2013)Razmara, Siahbani, Haffari, and
  Sarkar}]{razmara2013graph}
Majid Razmara, Maryam Siahbani, Reza Haffari, and Anoop Sarkar. 2013.
\newblock Graph {P}ropagation for {P}araphrasing {O}ut-of-{V}ocabulary {W}ords
  in {S}tatistical {M}achine {T}ranslation.
\newblock In \emph{Proc. of ACL}, pages 1105--1115.

\bibitem[{Saluja et~al.(2014)Saluja, Hassan, Toutanova, and
  Quirk}]{saluja2014graph}
Avneesh Saluja, Hany Hassan, Kristina Toutanova, and Chris Quirk. 2014.
\newblock Graph-based {S}emi-{S}upervised {L}earning of {T}ranslation {M}odels
  from {M}onolingual {D}ata.
\newblock In \emph{Proc. of ACL}, pages 676--686.

\bibitem[{Sennrich et~al.(2016)Sennrich, Haddow, and
  Birch}]{Sennrich2016NeuralMT}
Rico Sennrich, Barry Haddow, and Alexandra Birch. 2016.
\newblock {N}eural {M}achine {T}ranslation of {R}are {W}ords with {S}ubword
  {U}nits.
\newblock In \emph{Proc. of ACL}, pages 1715--1725.

\bibitem[{Sutskever et~al.(2014)Sutskever, Vinyals, and Le}]{SutskeverSeq2Seq}
Ilya Sutskever, Oriol Vinyals, and Quoc~V. Le. 2014.
\newblock {S}equence to {S}equence {L}earning with {N}eural {N}etworks.
\newblock In \emph{Proc. of NIPS}, pages 3104--3112.

\bibitem[{Tsvetkov and Dyer(2015)}]{tsvetkov2015lexicon}
Yulia Tsvetkov and Chris Dyer. 2015.
\newblock {L}exicon {S}tratification for {T}ranslating {O}ut-of-{V}ocabulary
  {W}ords.
\newblock In \emph{Proc. of ACL}, pages 125--131.

\bibitem[{Virpioja et~al.(2007)Virpioja, V{\"a}yrynen, Creutz, and
  Sadeniemi}]{Virpioja2007MorphologyAwareSM}
Sami Virpioja, Jaakko~J. V{\"a}yrynen, Mathias Creutz, and Markus Sadeniemi.
  2007.
\newblock {M}orphology-{A}ware {S}tatistical {M}achine {T}ranslation {B}ased on
  {M}orphs {I}nduced in an {U}nsupervised {M}anner.
\newblock In \emph{Proc. of the 11th Machine Translation Summit}.

\bibitem[{Wang et~al.(2012)Wang, Nakov, and Ng}]{wang2012source}
Pidong Wang, Preslav Nakov, and Hwee~Tou Ng. 2012.
\newblock Source {L}anguage {A}daptation for {R}esource-{P}oor {M}achine
  {T}ranslation.
\newblock In \emph{Proc. of EMNLP}, pages 286--296.

\bibitem[{Zhang et~al.(2014)Zhang, Liu, Li, Zhou, and
  Zong}]{zhang2014bilingually}
Jiajun Zhang, Shujie Liu, Mu~Li, Ming Zhou, and Chengqing Zong. 2014.
\newblock {B}ilingually-constrained {P}hrase {E}mbeddings for {M}achine
  {T}ranslation.
\newblock In \emph{Proc. of ACL}.

\bibitem[{Zhao et~al.(2015)Zhao, Hassan, and Auli}]{zhao2015learning}
Kai Zhao, Hany Hassan, and Michael Auli. 2015.
\newblock Learning {T}ranslation {M}odels from {M}onolingual {C}ontinuous
  {R}epresentations.
\newblock In \emph{Proc. of NAACL}, pages 1527--1536.

\bibitem[{Zou et~al.(2013)Zou, Socher, Cer, and Manning}]{zou2013bilingual}
Will~Y. Zou, Richard Socher, Daniel Cer, and Christopher~D Manning. 2013.
\newblock {B}ilingual {W}ord {E}mbeddings for {P}hrase-{B}ased {M}achine
  {T}ranslation.
\newblock In \emph{Proceedings of the 2013 Conference on Empirical Methods in
  Natural Language Processing}, pages 1393--1398.

\end{thebibliography}
\bibliographystyle{acl_natbib_nourl}

\clearpage

\begin{appendices}

\section{Validating Dataset Quality}
\label{appendix:dataset_validation}

To validate the quality of the automatically-constructed validation and test sets, we built an interface to enable native speakers to post-edit the generated translations. In this setup, speakers cannot provide their own translations for foreign words. Rather, they are shown a foreign sentence and its aligned English sentence, with the OOV and the translation respectively highlighted. They can edit the translation by modifying the highlighting on the English sentence. Speakers are allowed to highlight discontiguous spans.  For example, the translation of the Spanish word \textit{comer\'{e}}, as in ``No \textbf{comer\'{e}} la comida.'', would be \textit{will ... eat}, as in ``I \textbf{will} not \textbf{eat} the food''.

Volunteer native speakers validated the OOV datasets for  5 out of our 14 languages (Arabic, Bengali, Farsi, Russian, and Spanish). Many of the generated foreign OOVs and translations were not modified in the process, confirming their quality and the utility of the data-collection method.

\section{Number of Word Translation Pairs}
\label{appendix:seq2seq_oov_data_size}

\begin{table}[!htp]
\footnotesize
\centering
\begin{tabular}{lr}
\toprule
\begin{tabular}[c]{@{}l@{}}Source\\ Language\end{tabular} & \multicolumn{1}{l}{\begin{tabular}[c]{@{}l@{}}Number of \\ examples\end{tabular}} \\ 
\midrule
Amharic & 210.0K \tabularnewline
Arabic & 370.0K \tabularnewline
Bengali & 161.1K \tabularnewline
Farsi & 146.0K \tabularnewline
Hausa & 168.9K \tabularnewline
Hungarian & 938.3K \tabularnewline
Russian & 875.5K \tabularnewline
Somali & 179.5K \tabularnewline
Spanish & 944.1K \tabularnewline
Tamil & 54.6K \tabularnewline
Turkish & 349.9K \tabularnewline
Urdu & 123.9K \tabularnewline
Uzbek & 404.5K \tabularnewline
Yoruba & 233.8K \tabularnewline
\bottomrule
\end{tabular}
\caption{\label{tab:seq2seq_data} Number of word translation pairs (used to train the seq2seq OOV translator) for each language.}
\end{table}

\section{Seq2Seq OOV Translator Implementation Details}
\label{appendix:seq2seq_oov_details}

Our seq2seq models consist of 3-layer bidirectional LSTM networks with 1024 hidden units. After each LSTM layer except the last, we apply dropout of 0.3. Our character embeddings are 1024-dimensional. The model is trained with Adam \cite{Kingma2014AdamAM} with a constant learning rate of 0.0001 and a batch size of 128. The models are trained until sequence-level exact-match accuracy on the validation set shows no improvement for three epochs. We decode with a beam size of 1, and use the global attention with the general scoring function and input feeding as described in \newcite{Luong2015EffectiveAT}.

After training each model to convergence, we use the checkpoint with the highest exact match accuracy on a held-out validation set. Checkpoints are saved every 10,000 parameter updates and at the end of each epoch.

\section{Amount of MT Training Data For Each Language}
\label{appendix:sbmt_nmt_baselines_data_size}
\begin{table}[!htp]
\footnotesize
\centering
\begin{tabular}{lr}
\toprule
Source Language & Number of target tokens \\ 
\midrule
Amharic & 1.24M \tabularnewline
Arabic & 2.32M \tabularnewline
Bengali & 494.0K \tabularnewline
Farsi & 2.14M \tabularnewline
Hausa & 1.10M \tabularnewline
Hungarian & 5.20M \tabularnewline
Russian & 9.77M \tabularnewline
Somali & 1.40M \tabularnewline
Spanish & 11.90M \tabularnewline
Tamil & 262.5K \tabularnewline
Turkish & 2.23M \tabularnewline
Urdu & 527.1K \tabularnewline
Uzbek & 2.36M \tabularnewline
Yoruba & 1.11M \tabularnewline
\bottomrule
\end{tabular}
\caption{\label{tab:mt_data} Amount of training data (target-side tokens) used by SBMT and NMT systems for each language.}
\end{table}

\section{BPE NMT Baseline Implementation Details}
\label{appendix:nmt_baseline_details}
To train the BPE NMT models, we first apply byte pair encoding with 10K joins to the source and target data. We train a sequence-to-sequence model on the data with \texttt{OpenNMT-py} \citep{opennmt}, with git commit hash \texttt{0ecec8b}. The model is built and trained using the default hyperparameters: 2-layer LSTMs in both the encoder and decoder with 500-dimensional embedding vectors and RNN hidden states trained with SGD with an initial learning rate of 1.0. We edit the learning rate schedule from the default, training for 50 epochs and decaying after each epoch only when validation perplexity fails to increase. A checkpoint is saved after each epoch, and we use the checkpoint with the best validation perplexity to make test set predictions.
\end{appendices}
\end{document}